\documentclass[conference]{IEEEtran}
\IEEEoverridecommandlockouts
\usepackage{amsmath,amssymb,amsfonts}
\usepackage{algorithmic}
\usepackage{graphicx}
\usepackage{textcomp}
\usepackage{xcolor}
\def\BibTeX{{\rm B\kern-.05em{\sc i\kern-.025em b}\kern-.08em
    T\kern-.1667em\lower.7ex\hbox{E}\kern-.125emX}}

\usepackage{gensymb}  
\usepackage[numbers]{natbib}

\usepackage{algorithm}

\usepackage{graphicx, wrapfig, subcaption, setspace, booktabs}
\usepackage{dsfont}
\usepackage{url}
\usepackage{balance}

\begin{document}

\title{Graph Attention Network for Camera Relocalization on Dynamic Scenes
}

\author{\IEEEauthorblockN{Mohamed Amine Ouali}
\IEEEauthorblockA{\textit{Department of Computer Science} \\
\textit{University of Quebec at Montreal}\\
Montreal, QC, Canada \\
ouali.mohamed\_amine@courrier.uqam.ca}
\and
\IEEEauthorblockN{Mohamed Bouguessa}
\IEEEauthorblockA{\textit{Department of Computer Science} \\
\textit{University of Quebec at Montreal}\\
Montreal, QC, Canada \\
bouguessa.mohamed@uqam.ca}
\and
\IEEEauthorblockN{Riadh Ksantini}
\IEEEauthorblockA{\textit{Department of Computer Science} \\
\textit{University of Bahrain}\\
Zallaq, Bahrain \\
rksantini@uob.edu.bh}
}

\maketitle

\begin{abstract}
 We devise a graph attention network-based approach for learning a scene triangle mesh representation in order to estimate an image camera position in a dynamic environment. Previous approaches built a scene-dependent model that explicitly or implicitly embeds the structure of the scene. They use convolution neural networks or decision trees to establish 2D/3D-3D correspondences. Such a mapping overfits the target scene and does not generalize well to dynamic changes in the environment. Our work introduces a novel approach to solve the camera relocalization problem by using the available triangle mesh. Our 3D-3D matching framework consists of three blocks: (1) a graph neural network to compute the embedding of mesh vertices, (2) a convolution neural network to compute the embedding of grid cells defined on the RGB-D image, and (3) a neural network model to establish the correspondence between the two embeddings. These three components are trained end-to-end. To predict the final pose, we run the RANSAC algorithm to generate camera pose hypotheses, and we refine the prediction using the point-cloud representation.  Our approach significantly improves the camera pose accuracy of the state-of-the-art method from $0.358$ to $0.506$ on the RIO10 benchmark for dynamic indoor camera relocalization. 
\end{abstract}

\begin{IEEEkeywords}
Computer vision, Camera relocalization, 3D relocalization, Representation learning.\end{IEEEkeywords}

\section{Introduction}

The problem of relocalization aims to estimate the 6-degree-of-freedom camera pose (translation and rotation) in a known 3D environment. Several pioneering approaches have been proposed to solve this problem, and recent advancements have reached outstanding results, for instance, $99\%$ for $Pose(0.05m,5\degree)$ accuracy on static scenes \cite{bai2019survey}. The majority of these approaches are limited to static scenes mainly because the model is built to embed the 3D structure of the scenes, so any change in the environment will affect the dependency between the model and the scene. However, the real world is very dynamic. Changes like visual illumination and geometric changes are very frequent. This produces an urge to solve the more challenging problem of camera relocalization in dynamic scenes. \citet{wald2020beyond} showed that the astonishing results scored by the state-of-the-art method on static scenes turn out to be very poor when applied to dynamic scenes. The accuracy dropped from $99\%$ to drastically $28\%$. 

A recent work \cite{dong2021robust} proposed the NeuralRouting approach to solve the challenge of dynamic scenes. This approach inherits the decision tree framework of the state-of-the-art approaches for static scenes. The main idea is to build a binary decision tree that embeds physical knowledge of the environment. The tree is used to route points sampled from the example frame to the leaves, where 3D coordinates are estimated. \citet{dong2021robust} add an outlier-aware mechanism to their approach which rejects points that correspond to dynamic regions. Besides, they proposed an explicit hierarchical spacing of the scenes to associate the tree leaves with meaningful physical regions. These proposed techniques lead to an increase in the model accuracy on dynamic scenes. However, the NeuralRouting approach still suffers from some limits. First, there is the problem of point sampling: given the random aspect of the selected points, we could easily finish up with multiple points that belong to dynamic objects in the scene. Second, points belonging to dynamic regions could sometimes help identify the camera position, especially in configurations, where we lack texture information for static regions and the corresponding object was not moved from its place. Finally, models that embed the physical 3D knowledge of the scene could perform well in dynamic scenes if only static objects are considered for static scenes, whereas for dynamic scenes, this is only helpful for static unmoved objects. Therefore, a scene-independent model is a much better approach for targeting dynamic scenes.

\begin{figure}[t]\centering
\includegraphics[scale=0.3]{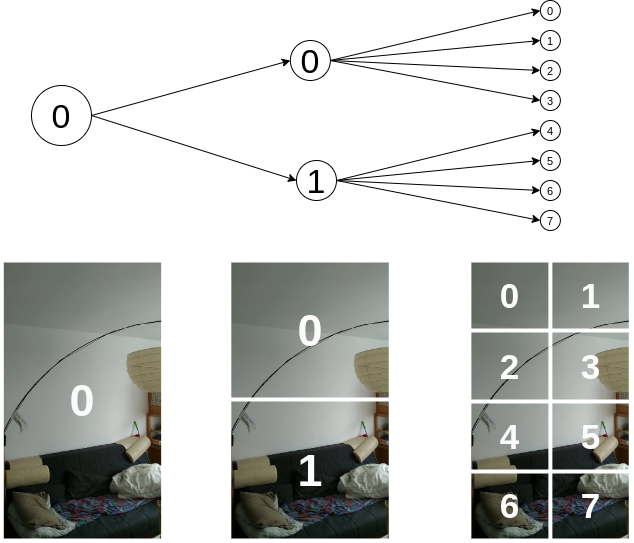}
\caption[]{Illustrating our hierarchical grid system for the query image. We show the three first level. At each level we split the current grids to subgrids. A vertex is routed through their corresponding grid of the image until reaching the final level, where we predict the offset of the vertex.}
\label{fig:grid_figure}
 \vspace{-15pt}
\end{figure}

To address the aforementioned observations, this paper presents a new approach (GATs-Loc) to solve the camera relocalization problem by leveraging the mesh representation of the scene and its semantic segmentation. First, we build a graph attention network model (GATs) to calculate the embedding of the triangle mesh vertices using their positions, normal vectors, colors, and semantic labels as input features, and the triangular polygon that forms the faces as edges. Second, we define a hierarchical grid configuration on the image as described by Figure \ref{fig:grid_figure}, and we design a CNN model to calculate the image cells embedding. Third, we apply a single layer neural network at each hierarchical level to match the embedding of the vertices with the embedding of the grid cells. If a vertex projection falls into a grid cell of the image, the grid cell is responsible for detecting the vertex. For the last hierarchical level, we predict the offset of the vertices in the grid cell. After establishing the correspondences between the RGB-D image pixels and the vertices of the 3D mesh, we apply the RANSAC algorithm with the Kabsch algorithm to predict the camera pose hypothesis. Finally, we refine the results with the Iterative Closest Point (ICP) algorithm using the mesh representation. 


    
    

\textbf{Contributions}. \textbf{(i)} We built a new scene-independent trainable network to establish 3D-3D correspondence between the mesh representation and the query image. \textbf{(ii)} Instead of relying on randomly sampled points or blind keypoints identification, we design a hierarchical grid system to establish 3D-3D correspondence. \textbf{(iii)} We evaluate GATs-Loc against the recently created benchmark RIO10 for camera relocalization in a dynamic indoor environment. Our experiments show that GATs-Loc outperforms the state-of-the-art approach in terms of camera $Pose(0.05m,5\degree)$ accuracy.


\section{Related work}


Early work used a simplistic but efficient approach to solve the camera relocalization problem. It consists of a database of images with known camera positions. In the prediction stage, the query image is compared to images in the database to find the most similar one. Then,  the new camera location is interpolated \cite{balntas2018relocnet, kim2017learned, arandjelovic2016netvlad}. Hierarchical localization \cite{middelberg2014scalable} uses the local feature to establish correspondence between the retrieved and query images. Then,  it calculates the query image camera pose using the ground-truth pose of the retrieved image and the respective correspondence. Using compact descriptors and advanced indexing techniques, these approaches can scale to bigger scenes \cite{kim2017learned, sattler2016large, torii201524}. Early image retrieval-based localization approaches have used hand-crafted methods to calculate global and local image descriptors. But, recent methods have used  CNNs to calculate these features.  Using compact descriptors and advanced indexing techniques, this approach can scale to larger scenes \cite{kim2017learned, sattler2016large, torii201524}. However, each component of these methods is trained separately, and there is a difficulty in matching the local features, especially in changing environments, because they are blindly calculated \cite{sattler2018benchmarking}. These methods also struggle if only a virtual 3D mesh representation is available. In such a case, the approach needs an efficient process to systematize the reference images of the dataset from the 3D mesh representation. Thus, the remaining question is how to generate such reference images, while keeping sufficient information for the pose prediction and without ending up with a lot of redundant images.

A more direct approach to solve the camera relocalization problem is the straight pose regression, formally named scene position regression \cite{acharya2019bim, Kendall_2017_CVPR, melekhov2017image}. The idea is to use a machine learning model trained with image and ground truth positions to derive the camera pose directly. This method only stores the model parameters instead of several images of the scene, so it reduces the time complexity which only depends on the model complexity. Furthermore, these approaches generalize well to new poses by creating an explicit or implicit internal representation of the 3D scene. CNNs are the main models used for camera pose regression.
This category of methods shows an increase in performance when given a sequence of consecutive frame images \cite{clark2017vidloc, saeedi2019characterizing}. The main problem with such methods is the lack of precision when feeding only one image.


The aforementioned methods do not exploit the available prior knowledge about 3D geometry. The perspective transformation exactly defines the relationship between the 3D position of the vertices, the image pixels, and the camera pose. To take advantage of such prior knowledge, structure-based or correspondence-based methods try to find the correspondence between the 2D/3D pixels in the image and 3D coordinates in the scene, by regressing the 3D coordinates and use these correspondences to estimate the pose. Such methods could be split into two steps. First, points' selection and scene coordinate regression. Second, camera pose estimation.

\begin{figure*}[!t]\centering
\begin{subfigure}{0.8\textwidth}
    \centering
    \includegraphics[width=\textwidth]{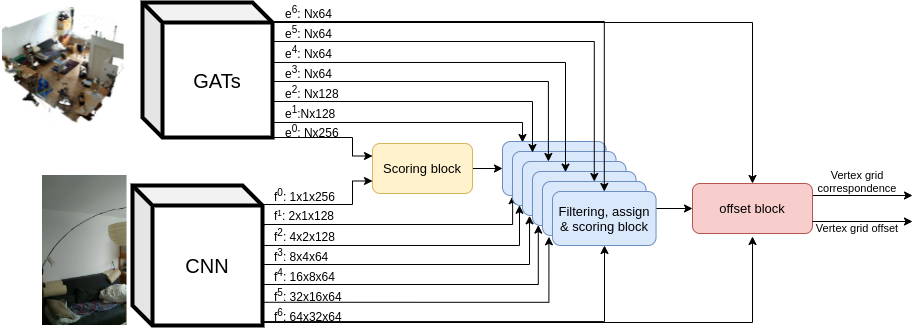}
    \caption{Global architecture.}
\end{subfigure}

\begin{subfigure}{.3\textwidth}
    \centering
    \includegraphics[width=\textwidth]{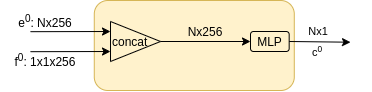}
    \caption{Scoring.}
\end{subfigure}%
\begin{subfigure}{.3\textwidth}
    \centering
    \includegraphics[width=\textwidth]{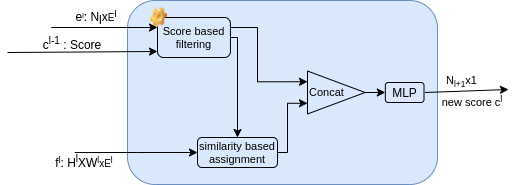}
    \caption{Filter and scoring.}
\end{subfigure}%
\begin{subfigure}{.3\textwidth}
    \centering
    \includegraphics[width=\textwidth]{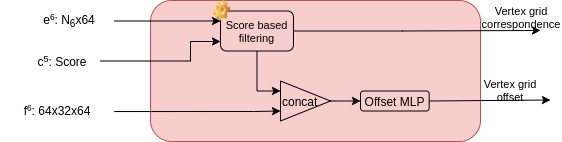}
    \caption{Filter and offset.}
\end{subfigure}%

\caption{The proposed GATs-Loc framework.}
\label{fig:overview}
\vspace{-10pt}
\end{figure*}

For the first step, the authors in \cite{brachmann2018learning} have used a CNN to learn the physical structure of the scene and regress the 3D coordinates from the query image. Other methods \cite{ cavallari2019let, golodetz2018collaborative, guzman2014multi} constructed a regression decision tree to establish the correspondence. First, they select points from the query frame using random sampling or keypoints identification. Second, they calculate the feature vector for these points used to build the tree and route the points to leaf nodes. Finally, the 3D coordinates are estimated for the points using a probability distribution identified in the training step. After establishing the point correspondence, the PnP (Perspective n Points) algorithm \cite{fischler1981random} for RGB images (2D to 3D correspondence) or the Kabsch algorithm \cite{kabsch1976solution} for RGB-D images (3D to 3D correspondence) are used to estimate the camera pose (the second step). The result is smoothed using the RANSAC algorithm \cite{fischler1981random}. It is an iterative method that calculates the parameters of a mathematical equation from the observed data that contains outliers. RANSAC is essential because some correspondence points are wrongly identified. 

Many variants of these approaches are then proposed. For example, \cite{brachmann2017dsac}  uses a CNN for 3D coordinates regression with a differential version of the RANSAC algorithm to enable end-to-end learning. \citet{cavallari2017fly} speed up the training for the regression tree on new scenes by only training the leaf nodes using a probability density. Finally, \citet{dong2021robust} proposed to adapt the tree-based method to the dynamic scene, by introducing an outlier-aware decision tree with neural routing, through the space partition of the scene, and achieved the best result on the RIO10 benchmark for camera relocalization in changing indoor environment.

\section{The Proposed GATs-Loc Approach}
Several camera relocalization methods are based on learning an implicit or explicit representation of the physical environment to regress the 3D coordinates. Such approaches are adequate for static scenes. However, a scene-independent model that does not encode the scene representation should be more robust to changes in the dynamic environment. This idea is partially introduced by the regression trees, where only the leaf nodes embed scene information to regress the 3D coordinates, and fully implemented by the image retrieval approach. We define our model as a scene-independent model. The general idea is to find the correspondence between the RGB-D image points and the vertices of the triangular mesh, and then use this correspondence to derive the camera's position.

Figure \ref{fig:overview} presents an overview of our approach. First, we hierarchically split the image into grids and define a single CNN model to calculate their embeddings (Section \ref{sec:imagegridsection}). Second, we define a graph attention network to find embeddings for the triangle mesh vertices (Section \ref{sec:GNN}). Third, we match the vertices with their corresponding grid regions (Section \ref{sec:matching}), and calculate their relative offset in the cell (Section \ref{sec:offset}). We calculate the exact 3D-3D  correspondence to infer the camera pose using these offsets. 

It is worth noting that, in our approach, we eliminate the step of random points sampling or keypoints identification used in other methods. Our method is similar to the image retrieval techniques, but it is more general as it uses the 3D mesh representation. Therefore, a lack of images for new poses does not affect the model and we do not rely on image interpolation to predict the camera pose.

\subsection{Image grid cells embedding}\label{sec:imagegridsection}

The first input to our model is the RGB-D image. We concatenate the depth map (representing the distance of each point from the camera) with the RGB image to obtain a four-channel 2D tensor $512\times288\times4$ used as an input to the CNN model. The image is then split into hierarchical regions as shown by Figure \ref{fig:grid_figure}. 
The hierarchical partition is similar to a tree representation. The root node of the tree represents the entire image. In the second level, we split the image with respect to its height to produce two equal cells (the height is approximately equal to two times the width). The same process is repeated for each subregion by splitting the regions into four parts. At the end, we obtain $64\times32$ regions corresponding to $8\times9$ pixels in the image. 


\begin{figure*}[!ht]
    \centering
    \includegraphics[width=0.85\textwidth]{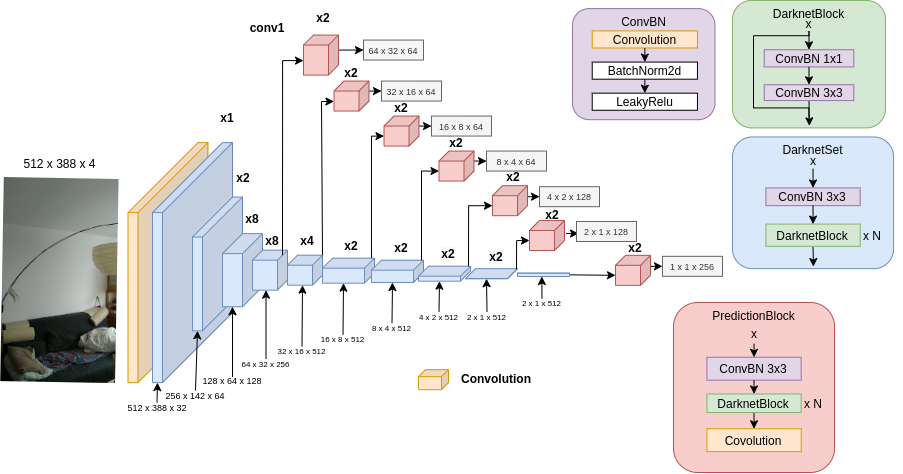}
    \caption{The global CNN architecture.}
    \label{fig:CNN}
    \vspace{-10pt}
\end{figure*}

We use the Darknet53 CNN model to calculate an embedding for each region in the image.  Darknet53 is defined for YOLOv3 \cite{redmon2018yolov3}, and it uses a succession of CNN blocks. Each time a CNN block is applied, it reduces the height and the width of the image by half and doubles the channel size. We use Darknet53 as the backbone for a larger model to calculate all image region embeddings in parallel.

Figure \ref{fig:CNN} illustrates the global CNN architecture. Our model uses a set of elementary building blocks. First, we have the ``ConvBN" building block consisting of a convolution layer followed by a batch normalization layer and a LeakyRelu. The ``DarknetBlock" building block consists of $1\times1$ ``ConvBN" followed by $3\times3$ ``ConvBN" blocks with residual connection. 
For the global architecture, we first apply 32 convolution filters of size $3\times3$ to the RGB-D image. Second, we apply a successive ``DarknetSet" block. Each time the ``DarknetSet" block set is applied, we use a $3\times3$ ``ConvBN" layer to divide the width and height by half and double the channel size. Then, we apply $N$ ``DarknetBlock" layers. In the second stage, we select the final and the previous six intermediate results. We apply a ``PredictionBlock" to produce seven tensors describing the image cells at each level for each tensor.

\subsection{ Triangle mesh vertices embedding}\label{sec:GNN}

 We use a graph neural network to calculate the embedding of the vertices of the triangle mesh. Such a representation (e.g., triangle mesh) can be created from image sequences using the ground truth camera poses. A triangle mesh is a set of vertices and faces that define a 3D shape. The vertices are defined by their positions and the faces define edges between three vertices. A 3D mesh is defined in a Euclidean 3D space and a trivial method to process such data structure using neural networks is voxelization. Voxelization discretizes the continuous 3D data to represent them in a 3D tensor. Depending on the discretization precision, the size of the grid can be very large, which could yield high unrealistic computations. To overcome such problems, we can keep the original sparse representation and use graph neural networks \cite{scarselli2008graph} to process such structure. Graph neural networks are used essentially for non-Euclidean data. However, they can also be useful for sparse Euclidean data, such as triangular mesh.
 
 \begin{figure*}[!t]
    \centering
    \includegraphics[width=0.95\textwidth]{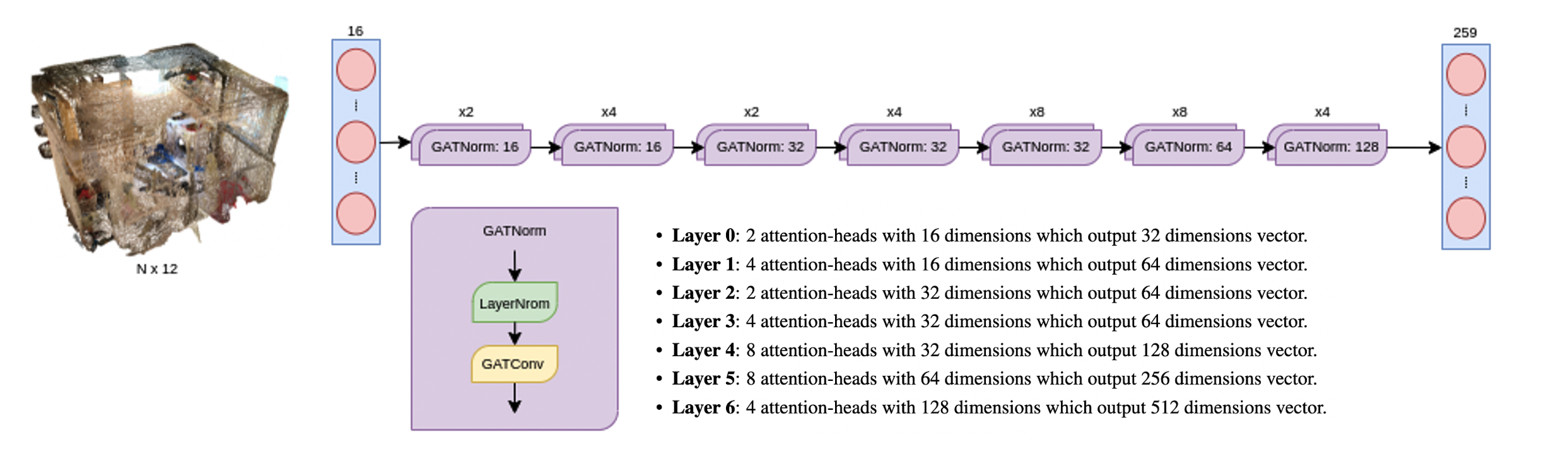}
    \caption{The global GNN architecture.}
    \label{fig:GNN}
     \vspace{-10pt}
\end{figure*}
 
 Our Graph Neural Network (GNN) is composed of seven layers. The number of layers represents how many indirect neighbor vertices to consider when calculating the vertex embedding. Our GNN model uses one type of layer which is the GATs \cite{velikovi2017graph} (Graph Attention Layer) defined by Eq. (\ref{eq:attention}). GAT uses the attention mechanism to calculate the weights needed to aggregate the neighboring vertices embedding, because some vertices could be more important than others. Eq. (\ref{eq:attention1}) applies a linear transformation to the previous layer vertices embedding using a learnable weight matrix. Eq. (\ref{eq:attention2}) calculates the attention scalar for all vertex pairs including the vertex itself (self-attention). In Eq. (\ref{eq:attention3}), we apply a softmax function to normalize the attention score of the node neighbors. Finally, in Eq. (\ref{eq:attention}) we aggregate the linear transformation using the attention scores and we apply the nonlinear activation function LeakyReLU.
 
 To improve the model capacity, we also leverage the multi-head attention mechanism. Eq. (\ref{eq:multi-head}) defines the multi-head attention in which we concatenate multiple embeddings calculated using different attention scores. $\alpha_{ij}^k$ represents the pair-wise attention score for the $k^{th}$ attention head. The linear transformation of the embedding $z_j^k$ is calculated using different learnable weight matrices, one for each attention head. The multi-head attention mechanism helps creating various embeddings, each focusing on a particular aspect of the 3D mesh representation that matches its specific hierarchical region. The output of the graph neural network is a descriptor vector for each vertex in the triangle mesh. The size of the descriptor vector is $768$ units which are divided into seven subvectors $256-128-128-64-64-64-64$ each of which is used with its corresponding hierarchical grid of the image.

\begin{align}
z_i^l &= W^{l}h_i^{l}\label{eq:attention1}\tag{\ref{eq:attention}.a}\\
a_{ij}^l &= LeakeyReLU( W_a^l (z_i^l || z_j^l))\label{eq:attention2}\tag{\ref{eq:attention}.b}\\
\alpha_{ij}^l &= \frac{e^{a_{ij}^l}}{\sum_{k \in N(i)\cup \{i\}} e^{a_{ik}^l}}\label{eq:attention3}\tag{\ref{eq:attention}.c}\\
h_i^{l+1} &= LeakeyReLU \Big(\textstyle\sum_{j \in N(i) \cup \{i\}} \alpha_{ij}^l z_j^l \Big)
\label{eq:attention}
\end{align}

\begin{equation}
\label{eq:multi-head}
h_i^{l+1} = \mathbin\Vert^K LeakeyReLU \Big(\textstyle\sum_{j \in N(i) \cup \{i\}} \alpha_{ij}^k z_j^k \Big)
\end{equation}

Figure \ref{fig:GNN} illustrates the global architecture of our graph neural network (GNN) model. The output of the proposed GNN model is a set of vertices with edges defined by the triangle mesh. Each vertex has a 12 dimension feature vector representing the $(x, y, z)$  position, the $(x_n, y_n, z_n)$ normal vector, $(R, G, B)$ color of the vertex, and  $(R_s, G_s, B_s)$ semantic segmentation which makes the difference between static and dynamic objects in the scene. First, we apply a single neural network layer to each feature vector to calculate the 16 embedding vectors. Then we apply seven ``GATNorm" blocks that include a normalization for the feature vector followed by a graph attention layer. The number of attentions-heads and dimensions of the resulting vector is different from one block to another. Each attention-head focuses on a set of neighboring vertices. The final layer of our model is a single neural network layer applied to each vertex descriptor to calculate an embedding vector of size 259. 




\subsection{Vertex-region matching}\label{sec:matching}

After calculating the embedding, we use a third neural network model to establish the matching between the vertices and the image zones. In the first level, the aim is to identify the vertices that belong to the image. To establish this, we calculate the absolute difference between the image embedding $f_j^l$ and the vertex embedding $e_i^l$. Then, we use a single-layer neural network to predict whether a vertex belongs to the image or not. Only the vertices that belong to the image are kept. In the next level, we assign the remaining vertices to regions of the image using distance similarity. Explicitly, we calculate the pairwise Euclidean distance between the regions and the vertices. Then, we match the vertices to the regions using the smallest distance. Next, we use a one-layer neural network to calculate the matching confidence to keep or discard the vertices.

Eq. (\ref{eq:confidence}) calculates the confidence of the assignment and it is applied at all levels. The confidence network is essential to discard vertices that are miss-assigned to regions. Eq. (\ref{eq:matching}) is used at all levels, except the first one, to establish correspondences. This process is repeated in all hierarchical splits of the image. Finally, the remaining vertices are assigned to $2,032$ regions defined by $64\times32$ grids of the image. The incremental hierarchical assignment (assigning the vertices to the region and then to its subregions) is used to reduce the computation complexity. The idea is to check only the sub-region of the region previously assigned to the vertex. For example, in level $i$, a vertex is assigned to region $B$ of the image. In the next level $i+1$, only the subregions of region $B$ are considered in the assignment.

\begin{align}
a_i^l &= \arg \min_{j}(distance(f_j^l,e_i^l) )   \label{eq:matching}\\
c_i^l &= sigmoid( W^l |f_{a^l}^l-e_i^l|+b^l)\label{eq:confidence}
\end{align}
where:
\par\vspace{\abovedisplayskip}\noindent\begin{tabular}{cl} 
$f_j^l$ & Embedding of image region $j$ at level $l$.  \\

$e_i^l$ & Embedding of vertex $i$ at level $l$.  \\

$W^l$, $b^l$ & Learnable weights at level $l$. \\

$a_i^l$ &  Region assigned to the vertex $i$ at level $l$. \\

$c_i^l$ & Probability of vertex $i$ assignment to region $a_i^l$.
  \end{tabular}\par\vspace{\belowdisplayskip}

\subsection{Relative offset}\label{sec:offset}

The previous step assigns the vertices to $2,032$ regions obtained by splitting the $512\times288$ image to $64\times32$ regions. Each region consists of $8\times9$ pixels. Therefore, we still need to calculate the exact location of the vertices in each region. Here, we do not use similarity-based method, but instead we adopt the offset estimation method, in order to reduce the computational complexity. Eq. (\ref{eq:offset}) represents the offset formula. We use a single layer perception to estimate $x,y\in [0,1]$ offset for each remaining vertex. First, we concatenate the embedding of the vertex and its corresponding last region. Then, we apply a neural network to calculate the coordinates.
\begin{equation}\label{eq:offset}
\begin{pmatrix}
x^l\\y^l
\end{pmatrix} = sigmoid( W^l (f_{a^l}^l || e_i^l)+b^l)
\end{equation}

\subsection{Loss functions}

To train the matching model end-to-end, we use four loss functions. Eq. (\ref{eq:loss_similarity}) defines the similarity loss function. This function aims to make the vertex embedding $e^l_{v}$ similar to the corresponding region embedding $f^l_{v}$. A trivial approach minimizes the Euclidean distance between the vertex embedding and its corresponding cell embedding. Although, a better approach is to use a loss function equivalent to triplet loss, in such an approach, we ensure that the Euclidean distance between the vertex embedding and its corresponding region embedding $d(e^l_v,f^l_{v})$ is smaller than the distance between the vertex and the other regions at the same level $d(e^l_v,f^l_r)$. The term $l_{e^l_v f^l_r}$ shows when $d(e^l_v,f^l_{v})$ is smaller than $d(e^l_v,f^l_r)$ the loss decreases and when $d(e^l_v,f^l_{v})$ is greater than $d(e^l_v,f^l_r)$ the loss increases. The margin between the two distance $margin^l$ depends on the level and it is chosen empirically with $margin^1=0.35$, $margin^2 = 0.3$, $margin^3 = 0.25$, $margin^4 = 0.2$, $margin^5 = 0.15$ and $margin^6 = 0.1$.

\begin{align}
L_s &=  \sum_{l=1}^6 \frac{1}{|V^l|} \sum_{v\in V^l} \sum_{r\in R^l} l_{e^l_v f^l_r} \label{eq:loss_similarity}\tag{\ref{eq:loss}.a}\\
L_{o} &=  \frac{1}{|V^6|} \sum_{v\in V^6} \mathds{1}_v [(x_v-\hat x_v)^2+(y_v-\hat y_v)^2]  \label{eq:loss_offset}\tag{\ref{eq:loss}.b}\\
L_c &= \sum_{l=0}^6 \frac{1}{|V^l|} \sum_{v\in V^l}   [-y^l_v log(p^l_v)+(1-y^l_v)log(1-p^l_v)] \label{eq:loss_confidence}\tag{\ref{eq:loss}.c}\\
L_n &= \sum_{l=0}^6 \frac{1}{|V^l|} \sum_{v\in V^l} ||e^l_v||+ \frac{1}{|R^l|} \sum_{r\in R^l} ||e^l_r||  \label{eq:loss_norm}\tag{\ref{eq:loss}.d}\\
L &= L_c + \lambda_s L_s+ \lambda_o L_{o}+ \lambda_n L_n \label{eq:loss}
\end{align}
where:
\par\vspace{\abovedisplayskip}\noindent\begin{tabular}{cl}

$l_{e^l_v f^l_r}$ &$=  max(0, (d(e^l_v,f^l_{v})-d(e^l_v,f^l_r)+margin^l))$.  \\

$V^l$ & Vertices set considered for level $l$.  \\

$\mathds{1}_v$ & $1$ if the vertex is assigned to its correct region, else $0$. \\

$R^l$ & Regions set for level $l$.  \\

$f^l_{v}$ & Embedding of vertex $v$ correct region at level $l$.  \\

$e^l_{v}$ & Embedding of the vertex at level $l$.  \\

$f^l_{r}$ & Embedding of region $r$ at level $l$.\\

\end{tabular}\par\vspace{\belowdisplayskip}

To reduce the time complexity, at level $i+1$, we only consider the subregions of the region assigned at level $i$. For example, if we consider a region $A$ at a level $i$ with four subregions $A.1$, $A.2$, $A.3$, and $A.4$ at the next level.  A vertex $v$ corresponds to the region $A.2$, then only the losses from the three other subregions $A.1$, $A.3$, and $A.4$ are considered in the loss function of vertex $v$. To compute the offset loss defined by Eq. (\ref{eq:loss_offset}), we use the known camera poses to calculate the ground truth offset. Then, we apply the mean square error to calculate the loss. The offset loss is calculated for the set of vertices that survived the routing process defined by $V^6$ and are assigned to the correct final cell. In Eq. (\ref{eq:loss_offset}), $\mathds{1}_v$ denotes if a vertex is assigned to its correct region at level 6. Eq. (\ref{eq:loss_confidence}) defines the confidence loss. To calculate it, we consider the vertices kept by the previous layers. Then, we use the cross-entropy loss function to calculate the loss, where $p^l_v$ defines the probability vertex $v$ assignment to the considered region at level $l$. $y^l_v$ is the ground truth probability which equals $1$ if the considered region is the correct region, otherwise $0$. Finally, Eq. (\ref{eq:loss_norm}) consists of minimizing the norm of the embedding for the vertices and the regions. This loss is a regularizer term that also helps to avoid gradient explosion. The global loss function  in Eq. (\ref{eq:loss}) is defined as a weighted sum of the other loss functions. In our experiments we set $\lambda_s=2$, $\lambda_o=15$ and $\lambda_n=0.2$.


After establishing the 3D-3D correspondence, we leverage the RANSAC \cite{fischler1981random} algorithm with the Kabsch \cite{kabsch1976solution} algorithm to predict the camera pose hypothesis, and we use the iterative closest point algorithm \cite{arun1987least} to refine the camera relocalization.





\section{Experimentation}

In this section, we present the results of our approach on RIO10 \cite{wald2020beyond}, a recently introduced benchmark for camera relocalization in dynamic indoors environments. The RIO10 dataset consists of 10 scenes. Each scene presents a real indoor environment with different configurations showing several geometric and visual changes encountered in a dynamic environment. The number of configurations varies from a scene to another. The first and second scene's configurations are used for training and validating the model, respectively. The remaining configurations are used to test the model. The ground truth poses are unknown for the test configurations and the results are obtained by submitting the prediction to the online platform: \url{http://vmnavab26.in.tum.de/RIO10/}.


\subsection{Implementation details}\label{sec:detail}

We used different configuration to train and test our GATs-Loc model. The common transformations are the resizing of the image to $512\times288$ pixels, and the normalization of the $x,\; y,\; z$ coordinates of the triangle mesh by centering the model to the world origin and scaling all coordinates by the same scalar to get all coordinates within $[-1,1]$. The same normalization parameters are used in the validation and testing process.

We trained the correspondence model using three stages. For the first stage,  we used the training configuration of all scenes for a better generalization of the correspondence model. We also applied data augmentation techniques to the image by applying blurring, Gaussian noise, and different illumination and contrast transformations. For the mesh, we applied random rotations within $[-180, 180]$ for the $z$-axis and within $[-5,5]$ for the $x$ and $y$-axis, which favors a rotation-invariant model. In the second stage, we do a transfer learning of the first stage model and we remove the data augmentation process for the triangular mesh. In the third stage, we refine the obtained model by using one scene configuration at one time. We assessed the refinement process using the validation configuration and it took between $1$ to $3$ epochs to refine the model to the target scene.

We used several techniques to validate and test our model. The first is to calculate the embedding of the triangle mesh vertices and retain it. Since the vertices embeddings do not depend on the image frame being evaluated, the graph neural network model can be removed by memorizing the embeddings to reduce the time complexity. The second technique is multi-routing throw the hierarchical level. Due to the potential error in assigning the vertices to the correct region of the image, we allow multi region matching to a single vertex in the test phase. For our configuration, we allow one assignment for the second hierarchical level, so the vertex is only matched with the top or the bottom half of the image. Then we keep the best $3$, $3$, $3$, and $4$ image regions matching for the third, fourth, fifth, and sixth levels, respectively. Concretely, we allow each vertex to be assigned to three grid cells in the third, fourth, and fifth level, and to four grid cells in the sixth hierarchical level. The considered grids are selected based on the minimum embedding distance. Finally, we get four potential 3D correspondence for each vertex that survived the routing process. All assignments are used in the camera pose prediction. Our experimentation shows that this trick can help correct many points' correspondences, especially points near the split boundary.

\begin{algorithm}[t]
\caption{Custom RANSAC algorithm for pose estimation.}
\label{alg:algorithm}
\textbf{Input}: 3D-3D correspondence from the global matching model.\\
\textbf{Output}: $4 \times 4$ camera transformation matrix.~\\
\begin{algorithmic}[1]
\STATE Let $it=0$
\STATE Let $H=[]$ Pose hypothesis table
\WHILE{$it<1024$ }
\STATE Select 3 points pair $A, B, C$ from the 3D-3D correspondence set.
\IF {farEnough(A, B, C) and rigidTransformation(A, B, C)}
\STATE $M_t$ = Kabsch(A, B, C).
\STATE Append $M_t$ to the hypothesis list $H$.
\ENDIF
\STATE $it += 1$
\ENDWHILE
\STATE Calculate a score for each camera pose hypothesis in $H$ using the vertices of static objects.

\STATE $M$ = Select the best camera hypothesis using the previous calculated score.
\STATE Refine the camera pose $M$ using ICP with static points.
\STATE \textbf{return} $M$.
\end{algorithmic}
\label{alg:pose}
\end{algorithm}

 Algorithm \ref{alg:pose} illustrates the steps needed to predict the camera pose from the 3D-3D correspondence. We start by generating 1024 camera pose hypotheses that satisfy two conditions. The ”farEnough” condition ensures that the selected vertices' pairs are far from each other. The "rigidTransformation" condition ensures that the distances between the points in the world space and the camera space are similar enough. We rank them after generating the camera hypothesis and select the best-score hypothesis. To calculate the score, we take the 3D points identified in the camera space and multiply them with the inverse camera hypothesis matrix to find their coordinates in the world space. Then, we calculate the mean square error between the calculated world coordinates and those identified in the matching process. Only points corresponding to static objects are used to calculate the mean square error. In the end, we take the inverse of the mean square error.

\subsection{Correspondence model results}\label{sec:eval1}

\begin{figure}[t]\centering
\includegraphics[scale=0.3]{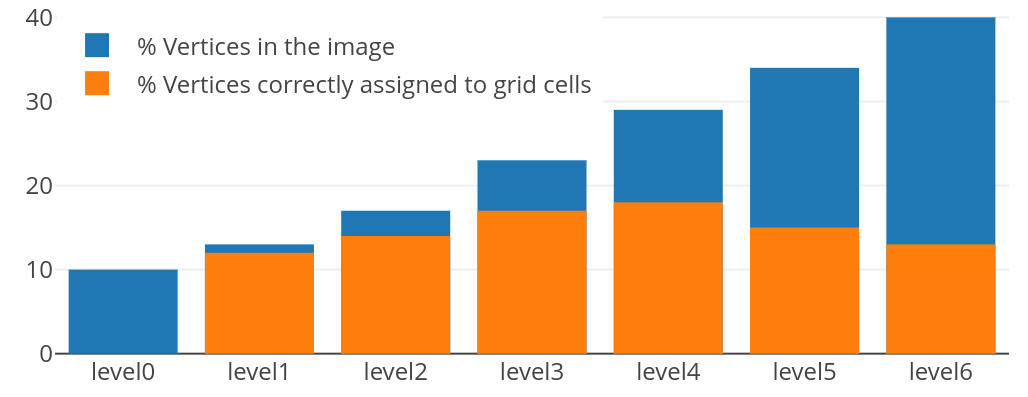}
\caption[]{The rates of the correctly predicted vertices belonging to the image along with the rates of their assignment for seq01\_02.}
\label{fig:rates}
 \vspace{-10pt}
\end{figure}

In this experiment, we evaluate the hierarchical correspondence model. Figure \ref{fig:rates} illustrates the results of the validation configuration of the first scene (seq01\_02). The figure shows the mean results for $2,069$ images. Therefore, the results can differ from one image to another. To better understand the results, we note that the number of vertices for this scene is $55,895$ and the mean number of vertices corresponding to each image is approximately $1,000$, which implies that a random selection of vertices will have an accuracy of $1.7\%$. As shown in the figure, the first level can correctly identify $10\%$ of the vertices belonging to the image and this percentage keeps increasing at each level until reaching the $6^{th}$ level, where about $40\%$ of the identified vertices belong to the image. The orange color in Figure \ref{fig:rates} represents the rate of the vertices that were correctly assigned to their respective grids. Their rate keeps on increasing until reaching level four, and then the rate decreases. This pattern is comprehensible. As we split the image further in the hierarchical levels, the neighboring cells embedding become more similar, which results in wrong correspondence. Such a deviation in identifying the correspondence has small to no effect because it is corrected by the multi-label assignment presented in Section \ref{sec:detail}, which makes it less significant.

\subsection{Main results}\label{eval1}

\begin{table*}
  \caption{Results on the RIO10 benchmark. The bold and blue rank the first and second metric, respectively.}
\label{tab:rio10_full_result}

\begin{center}

  \begin{tabular}{lrrrrrr}
\toprule
Method  & Score & DCRE(0.05) & DCRE(0.15) & Pose(0.05m,5\degree) &  Outlier(0.5) & NaN \\
\midrule
GATs-Loc (RGB-D) & \color{blue} 1.401&	\color{blue} 0.533	& 0.579&	\textbf{0.506}&	0.132&	0.196	\\
\hline
\small NeuralRouting \cite{dong2021robust} (RGB-D) & \textbf{1.441} &	\textbf{0.538} &	\textbf{0.615} &	\color{blue}0.358 & \color{blue} 0.097 & 0.227  \\
\small Grove v2  \cite{cavallari2019real} (RGB-D) & 1.162 & 0.416 & 0.488 & 0.274 & 0.254 & 0.162 \\
 \small Grove  \cite{cavallari2017fly} (RGB-D)  & 1.240 & 0.342 & 0.392 & 0.230 & 0.102 & 0.452  \\
\small KAPTURE  \cite{humenberger2020robust}(RGB)  & 1.338 & 0.447 & \color{blue} 0.612 & 0.176 & 0.109 & \textbf{0.000}  \\
 \small D2-Net \cite{Dusmanu2019CVPR}(RGB)  & 1.247 & 0.392 & 0.521 & 0.155 & 0.144 & 0.014  \\
\small HF-Net (trained) \cite{sarlin2019coarse}(RGB)  & 0.789 & 0.192 & 0.300 & 0.073 & 0.403 & \textbf{0.000} \\
\small Active Search \cite{sattler2016efficient}(RGB)  & 1.166 & 0.185 & 0.250 & 0.070 & \textbf{0.019} & 0.690  \\
\small HF-Net  \cite{sarlin2019coarse}(RGB)  & 0.373 & 0.064 & 0.103 & 0.018 & 0.690 & \textbf{0.000}  \\
\small NetVLAD  \cite{arandjelovic2016netvlad}(RGB)   & 0.507 & 0.008 & 0.136 & 0.000 & 0.501 & \color{blue} 0.006  \\
\small DenseVLAD  \cite{torii201524} (RGB)  & 0.575 & 0.007 & 0.137 & 0.000 & 0.431 & \textbf{0.000}	\\

\bottomrule
\end{tabular}
\end{center}
\end{table*}

Table \ref{tab:rio10_full_result} illustrates the  quantitative results of our framework and other state-of-the-art methods for camera relocalization on the RIO10 test set. In indoor camera relocalization, camera pose accuracy Pose(0.05m,5\degree) is regarded as the more common and direct measurement for assessing the results \cite{dong2021robust, tang2021learning}. This metric represents the rate of test frames whose rotation and translation errors are within 5 degrees and 5cm, respectively. For the Pose(0.05m,5\degree) metric. Our result (0.506) surpasses the state-of-the-art approach (0.358) by about 15\%. As shown in Table \ref{tab:rio10_full_result}, there is a correlation between the DCRE(0.05) and the Pose(0.05m,5\degree). In fact, both metrics are proportional. Test frames considered by Pose(0.05m,5\degree) metric are  considered by the DCRE(0.05). Similar prediction and ground truth camera pose will always show similar visual perception (high DCRE value). Regarding DCRE(0.05) metric, the results show that our framework is the second-best $0.533$ compared to NeuralRouting $0.538$. GATs-Loc has similar results for DCRE(0.05) and Pose(0.05m,5\degree) in contrast to NeuralRouting approach, where the DCRE(0.05) is much better than the Pose(0.05m,5\degree) error. 

The table also covers four more metrics. The NaN metric corresponds to the percentage of test frames, where the model could not find the corresponding camera pose, which  is about 19\% of the test frames for our model. The missing camera pose predictions are the results of insufficient 3D-3D corresponding. In other words, not enough vertices survived the routing process. The Outlier(0.5) and the DCRE(0.15) are related to the DCRE metric. DCRE(0.15) corresponds to the percentage of test frames with a DCRE value within 0.15. The Outlier(0.5) metric corresponds to the test frames rate with DCRE greater than 0.5. Our approach shows less competitive results for Outlier(0.5) and the DCRE(0.15) metric compared to the state-of-the-art. The  Score metric (first column of the table) also depends on the DCRE metric as it equals $1+DCRE(0.05-Outlier(0.5))$. Our framework is second to NeuralRouting if we consider the Score metric, because it does not depend on the Pose(0.05m,5\degree) metric. We show the qualitative results of the matching model in the appendix.

\section{Ablation study}

To demonstrate the influence of various design choices, we performed an ablation study on the RIO10 validation dataset\footnote{We use the validation dataset because the ground-truth labels of the test dataset are not publicly available. In addition, the online evaluation platform imposes a two-week duration between submissions to forbid parameter tuning on the test dataset.}. The results of our experiments are shown in Table \ref{tab:ablation}.

\textbf{GCNs instead of GATs}. This experimental study shows the importance of using graph attention layers in comparison to a simple graph convolution network. Specifically, it demonstrates the importance of the multi-head attention mechanism used by the graph neural network layers. In fact, the graph convolution layer traits the vertex and its neighbors in the same manner when aggregating their results. However, not all vertices should be traits similarly in the mesh representation because some vertices are more critical than others. For example, vertices representing the corners or the borders of an object are more important than those located in a flat area. Hence, the attention mechanism is critical when calculating the embedding of the vertices. The attention mechanism performs an intelligent aggregation based on the calculated weight for each adjacent vertex.  

\textbf{Three graph attention layers instead of seven}. For the graph neural network model,  a three-layer graph neural network will aggregate the vertices representation that is at most two vertices units further from the vertex. By increasing the number of layers, we consider more vertices that are further away from the current vertex when calculating the embedding. The experiment shows a drop in performance when we decrease the number of layers. Further experiments show that the model became incapable of associating the correct vertices to the image at the first hierarchical level. By considering less distant vertices, the calculating embedding becomes a local representation that is harder to match with the global representation of the image.

\begin{table}
  \caption{Ablation study for the RIO10 validation dataset.}
  \label{tab:ablation}
  \begin{tabular}{lc}
    \toprule
     & Pose(0.05m,5\degree)\\
    \midrule
    GCNs instead of GATs & 28.12\% \\
    Three GATs layer instead of seven & 27.39\%  \\
    GNN without vertices semantic & 39.20\% \\
    GNN without vertices normal & 50.18\% \\
    GNN without vertices colors & 19.66\% \\
    GNN without vertices position & 25.32\% \\
    GATs-Loc & \textbf{53.93\%} \\
  \bottomrule
\end{tabular}
\end{table}

\textbf{Removing the vertices semantic/normal/colors/position}. In this experiment, we wanted to study the effect of reducing the information feed to our graph neural network model. Each time, we remove a single input data and make comparisons on the RIO10 validation dataset. The most important information seems to be the colors. Obviously, the network relies too much on the colors to match the image and the mesh vertices. In fact, this information exists in both data, and it is very easy for the model to make the association between them. The result also shows that the position of the vertices is also very significant to achieve the final results. Theoretically, the vertex position is important because it helps identify the object shapes, which can then be matched to the similar shape identified visually in the image. The network seems to also rely on semantic information. In computer vision, CNNs have proven their ability to identify the classes of objects present in the scene. Therefore, our CNN model can identify the class of static objects in the images and then match them with the semantic information available in the mesh representation. As for the vertices normals, we notice a small performance drop in Pose(0.05m,5\degree) accuracy compared to the other input information. When normalizing the data, we have centered the model on the origin of the 3D world. Consequently, the model could deduce the normal information for some vertices.

\section{Conclusion}


We have designed a triangle mesh-based camera relocalization framework for dynamic indoor environments, in which we leverage the learning capability of graph neural networks. We also proposed a hierarchical structure of the image to identify the 3D-3D correspondence between the test frame and the vertices of the triangle mesh, without randomly sampled points or blind keypoints identification. Finally, we show the performance of our results by validating our method on the RIO10 benchmark. Our framework proposes a novel way for solving the camera relocalization problem, and different ideas can be proposed to improve the overall performance. For instance, in our ongoing work, we plan to extend our approach to RGB-only test images, which are based on more common and less expensive hardware compared to RGB-D images. We are investigating better representation for the semantics which will be based on the probability that an object changes its location in the scene. This could significantly improve the performance of the model. We also plan to take advantage of Vision Transformers (ViT) to calculate image region embedding instead of CNN. Finally, we intend to build a hierarchical graph neural network model in which we compact the number of vertices after each applied layer.

\balance

\bibliographystyle{ieeetranN}
\scriptsize{
\bibliography{main_references}
}

\onecolumn

\appendix
\section*{Matching model results} \label{sec:reproduction}

\begin{figure*}[!ht]\centering

\begin{subfigure}{.18\textwidth}
    \centering
    \includegraphics[width=\textwidth]{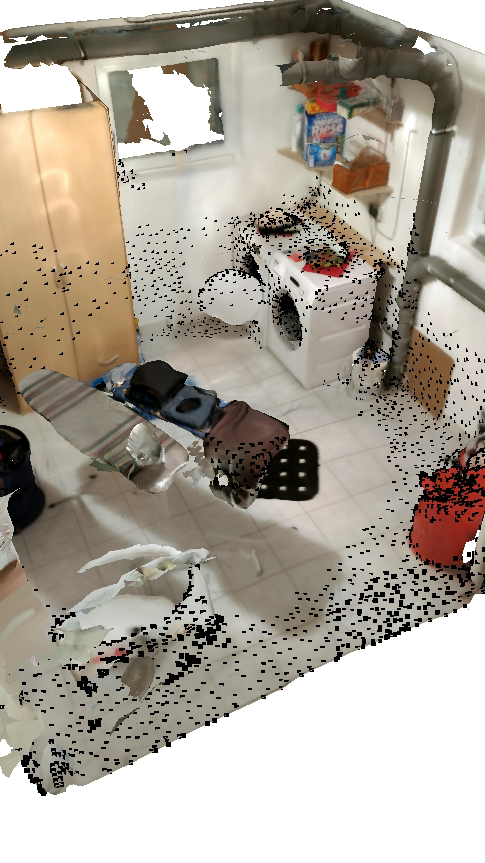}
    \caption{Level 1}
\end{subfigure}%
\hspace{.01\textwidth}
\begin{subfigure}{.18\textwidth}
    \centering
    \includegraphics[width=\textwidth]{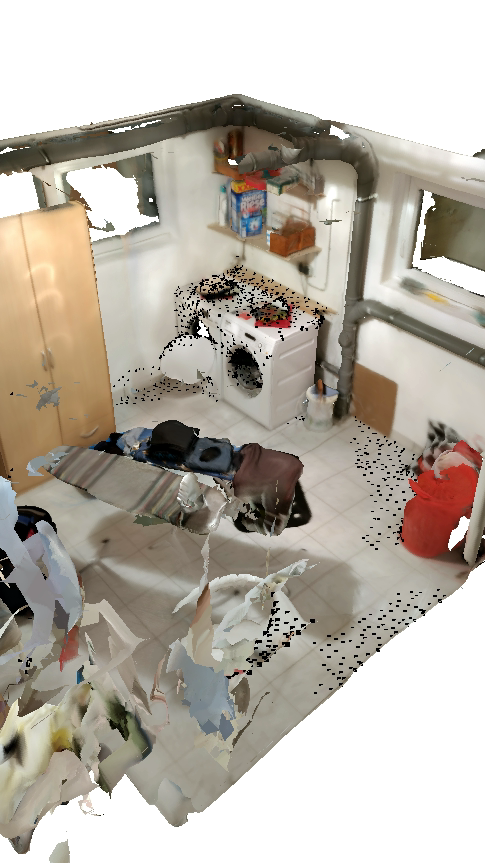}
    \caption{Level 2}
\end{subfigure}%
\hspace{.01\textwidth}
\begin{subfigure}{.18\textwidth}
    \centering
    \includegraphics[width=\textwidth]{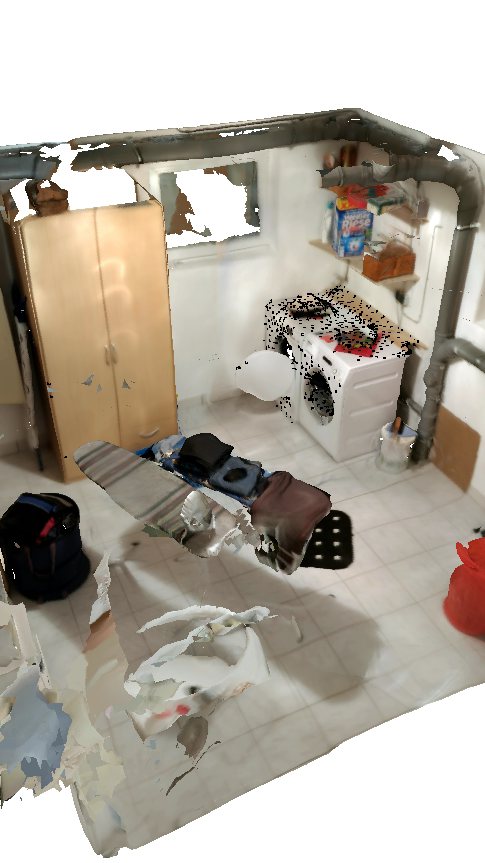}
    \caption{Level 4}
\end{subfigure}%
\hspace{.01\textwidth}
\begin{subfigure}{.18\textwidth}
    \centering
    \includegraphics[width=\textwidth]{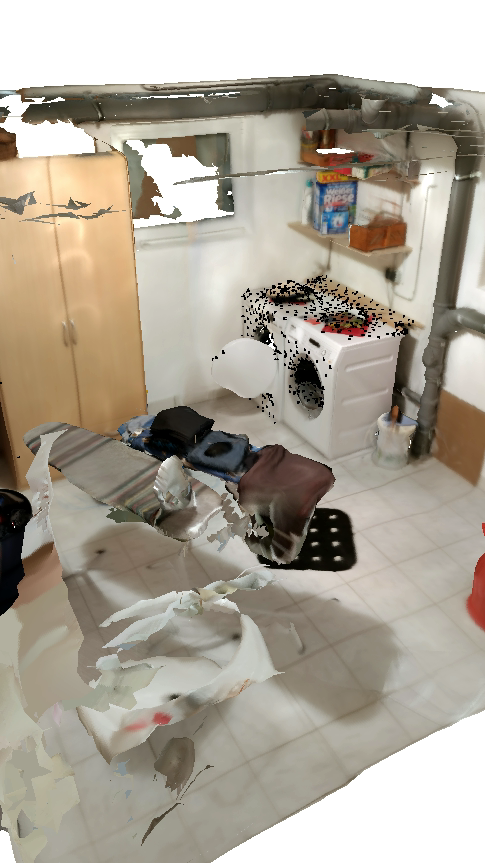}
    \caption{Level 6}
\end{subfigure}%
\hspace{.01\textwidth}
\begin{subfigure}{.18\textwidth}
    \centering
    \includegraphics[width=\textwidth]{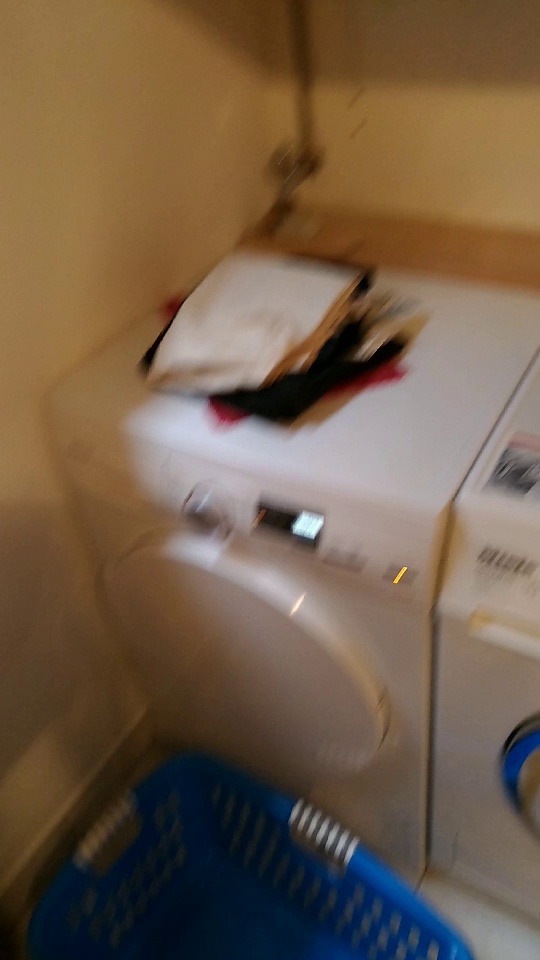}
    \caption{Test frame} \label{fig:original}
\end{subfigure}%

\caption{An example showing the vertices that survived the routing process (in black) throw level 1 (Figure a), level 2 (Figure b), level 4 (Figure c), and level 6 (Figure d) for a specific test frame which is represented by Figure e. In this example, we illustrates some illumination and geometric changes between the triangle mesh and the test frame. We can perceive that the door of the washing machine is closed in the test frame (Figure e) and the blue basket was moved. The test frame also illustrate illumination change. }
\label{fig:example_levels}
\end{figure*}

\begin{figure*}[!ht]\centering

\begin{subfigure}{.18\textwidth}
    \centering
    \includegraphics[width=\textwidth]{figures/results/level6.png}
\end{subfigure}%
\hspace{.01\textwidth}
\begin{subfigure}{.18\textwidth}
    \centering
    \includegraphics[width=\textwidth]{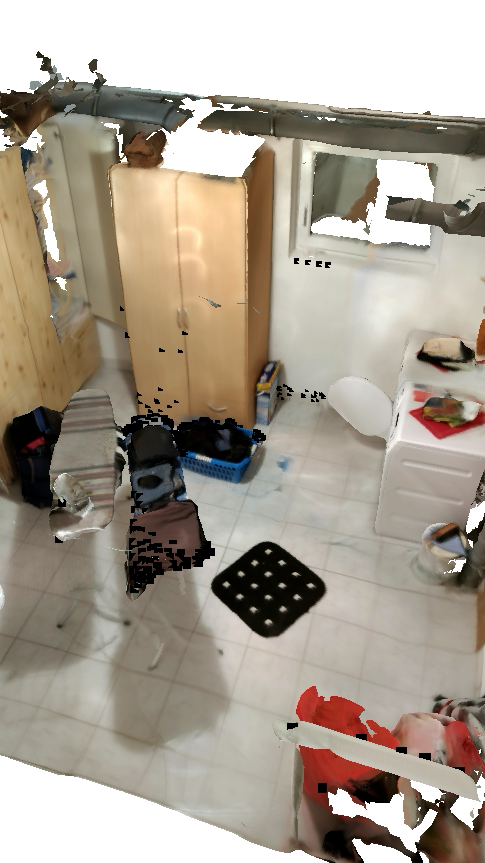}

\end{subfigure}%
\hspace{.01\textwidth}
\begin{subfigure}{.18\textwidth}
    \centering
    \includegraphics[width=\textwidth]{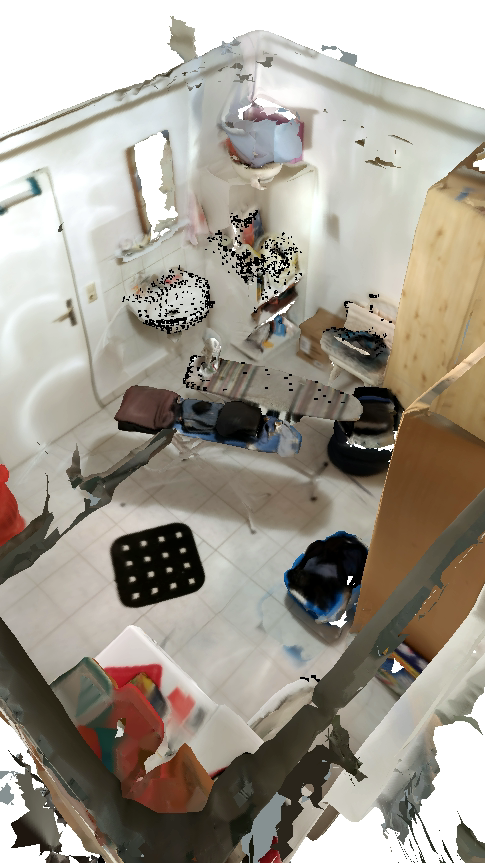}

\end{subfigure}%

\vspace{.01\textwidth}

\begin{subfigure}{.18\textwidth}
    \centering
    \includegraphics[width=\textwidth]{figures/results/original.jpg}

\end{subfigure}%
\hspace{.01\textwidth}
\begin{subfigure}{.18\textwidth}
    \centering
    \includegraphics[width=\textwidth]{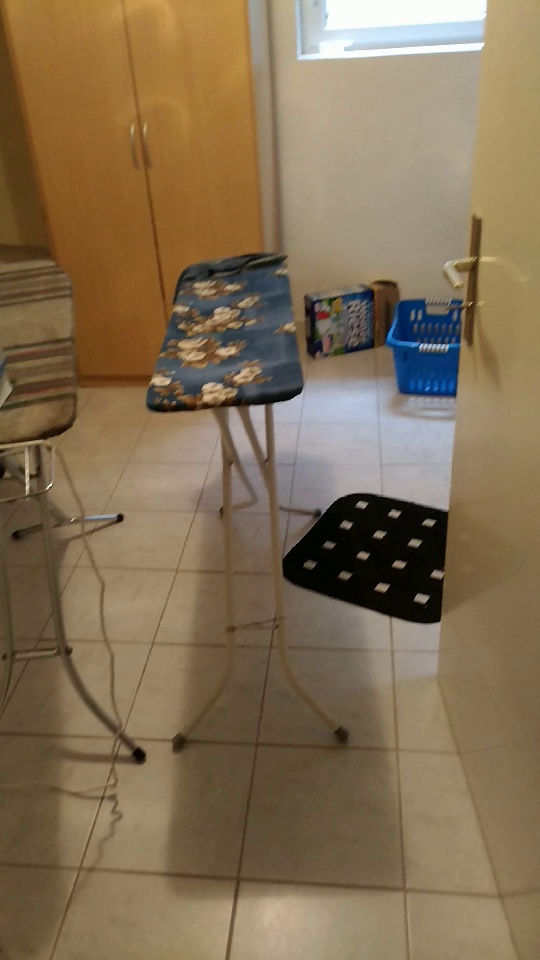}

\end{subfigure}%
\hspace{.01\textwidth}
\begin{subfigure}{.18\textwidth}
    \centering
    \includegraphics[width=\textwidth]{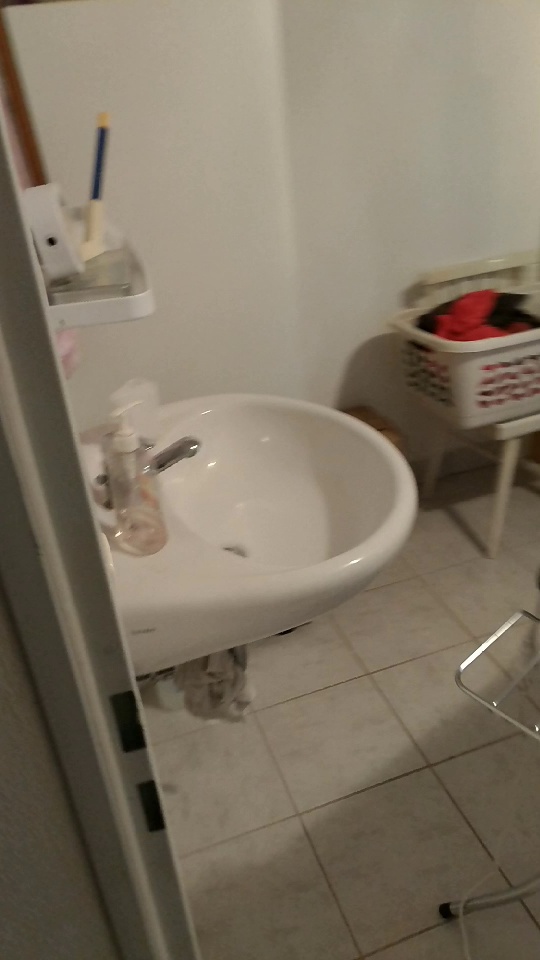}

\end{subfigure}%

\caption{We show the vertices that survived the routing process (in black) until level 6 for 3 test frames. }
\label{fig:example_levels_result}
\end{figure*}

\end{document}